\documentclass{IOS-Book-Article}

\usepackage{mathptmx}
\usepackage{soul}\setuldepth{article}
\usepackage{enumitem}
\usepackage[numbers,square,sort&compress]{natbib}
\usepackage{booktabs}
\usepackage{graphicx}
\usepackage{float}
\usepackage{threeparttable}
\usepackage{times}
\normalfont
\usepackage[T1]{fontenc}
\def\hb{\hbox to 11.5 cm{}}
\begin{document}
\pagestyle{headings}
\def\thepage{}
\begin{frontmatter}

\title{Automating SPARQL Query Translations between DBpedia and Wikidata}

\markboth{}{July 2025\hb}

\author[A]{\fnms{Malte Christian} \snm{Bartels}\orcid{0009-0006-2113-3322}%
\thanks{Corresponding Author: Malte Christian Bartels, Malte.C.Bartels@stud.leuphana.de}},
\author[A]{\fnms{Debayan} \snm{Banerjee}\orcid{0000-0001-7626-8888}}
and
\author[A]{\fnms{Ricardo} \snm{Usbeck}\orcid{0000-0002-0191-7211}}

\runningauthor{Bartels et al.}
\address[A]{Leuphana University of Lüneburg, Lüneburg, Germany}

\begin{abstract}
\textit{Purpose}: This paper investigates whether state-of-the-art Large Language Models (LLMs) can automatically translate SPARQL between popular Knowledge Graph (KG) schemas. We focus on translations between the DBpedia and Wikidata KG, and later on DBLP and OpenAlex KG. This study addresses a notable gap in KG interoperability research by evaluating LLM performance on SPARQL-to-SPARQL translation. \\
\textit{Methodology}: Two benchmarks are assembled, where the first aligns 100 DBpedia–Wikidata queries from QALD-9-Plus dataset; the second contains 100 DBLP queries aligned to OpenAlex, testing generalizability beyond encyclopaedic KGs. Three open LLMs: Llama-3-8B, DeepSeek-R1-Distill-Llama-70B, and Mistral-Large-Instruct-2407 are selected based on their sizes and architectures and tested with zero-shot, few-shot, and two chain-of-thought variants. Outputs were compared with gold-standard answers, and resulting errors were systematically categorized. \\
\textit{Findings}: We find that the performance varies markedly across models and prompting strategies, and that translations for Wikidata to DBpedia work far better than translations for DBpedia to Wikidata. The largest model, Mistral-Large-Instruct-2407, achieved the highest accuracy, reaching 86\% on the Wikidata $\rightarrow$ DBpedia task using a Chain-of-Thought approach. This performance was replicated in the DBLP $\rightarrow$ OpenAlex generalization task, which achieved similar results with a few-shot setup, underscoring the critical role of in-context examples. \\ 
\textit{Value}: This study demonstrates a viable and scalable pathway toward KG interoperability by using LLMs with structured prompting and explicit schema-mapping tables to translate queries across heterogeneous KGs. The method's strong performance when applied to general purpose KGs and specialized scholarly domain suggests its potential as a promising approach to reduce the manual effort required for cross-KG data integration and analysis. 
\end{abstract}

\end{frontmatter}
\markboth{July 2025\hb}{July 2025\hb}

\newpage

\section{Introduction}\label{sec:introduction}

KGs like Wikidata~\citep{Vrandecic.2014} and DBpedia~\citep{Lehmann.2015} represent vast stores of interconnected facts, typically structured as subject-predicate-object triples~\citep{Zeng.2021} and often encoded using the Resource Description Framework (RDF)~\citep{Schreiber.2014}. This machine-readable graph structure, queried via the SPARQL protocol~\citep{Jorge.2009, Ali.2022}, offers significant value for semantic web technologies and artificial intelligence (AI) by improving information accessibility and reusability~\citep{Hogan.2022}. However, the true potential of combining insights across these rich repositories is often hindered by fundamental interoperability challenges. A prime example of this challenge lies in the portability of queries; SPARQL, the standard language for querying KGs~\citep{Ali.2022}, is tightly coupled to individual KG schemas. Consequently, a query crafted for one KG, like DBpedia, will rarely function correctly on another, such as Wikidata, without substantial manual adaptation due to differing predicates, classes, or entity identifiers~\citep{Hofer.2024}. This lack of query portability is a critical bottleneck to seamless knowledge integration.

Thus, automating the translation of SPARQL queries between different KGs is a crucial step towards unlocking true interoperability. Such automation would empower users and applications to seamlessly query, integrate, and cross-validate information across multiple KGs, thereby broadening access to verified data and enhancing the reliability of query results. This also unlocks the portability of current Knowledge Graph Question Answering (KGQA) datasets, which may have been created for a single KG.  Recent advancements in AI, particularly with Large Language Models (LLMs) - powerful systems typically based on the transformer architecture~\citep{Ashish.2017} — present a promising avenue for this complex translation task. Given their demonstrated capabilities in understanding complex patterns and generating structured text like code~\citep{Brown.2020}, and their established proficiency in composing SPARQL queries from natural language questions~\citep{Yin.2021}, LLMs are compelling candidates for transforming SPARQL queries between disparate KG schemas.

This automated SPARQL-to-SPARQL translation capability is not merely a technical convenience; it is foundational for realizing the full potential of synergistic LLM and KG integration. While LLMs offer powerful generative capabilities, they are also prone to "hallucinations", generating plausible yet incorrect information, and can perpetuate biases~\citep{Agrawal.2024}. Integrating LLMs with verifiable external KGs offers a path to mitigate these limitations by grounding their outputs in structured, reliable facts~\citep{Pan.2024}. Moreover, KGs can support complex reasoning tasks for LLMs, enabling them to decompose broad questions into precise sub-queries over graph structures~\citep{Liang.2024}. For LLMs to effectively leverage the rich and diverse landscape of existing KGs, rather than being confined to a single KG's schema, they must be able to interact fluently across these varied structures. Automated query translation thus serves as the bridge, enabling LLMs to query, reason across, and harness the combined strengths of multiple, heterogeneous KGs. Therefore, the development of effective methods for SPARQL-to-SPARQL translation can contribute significantly to advancing KG interoperability and enhancing the capabilities of KG-aware AI systems.

\noindent Motivated by this need, this study investigates the challenge of cross-KG interoperability through the lens of automated SPARQL query translation. It focuses on developing and evaluating LLM-based methods to translate SPARQL queries between DBpedia and Wikidata, two widely used yet structurally distinct KGs. The QALD-9-plus dataset~\citep{Perevalov.2022}, providing aligned natural language questions (NLQs) and SPARQL queries for both KGs, serves as a primary resource. Furthermore, the potential for generalization of promising methods is examined by applying them to a different pair of KGs in the scholarly domain: DBLP~\citep{Ley.2002} and OpenAlex~\citep{Priem.2022}, to verify their performance on a non-encyclopaedic use case.

\noindent The overall objective is to advance KG interoperability, thereby simplifying access to reliable, structured data across diverse platforms. Our main contribution is to analyse the performance of different open-source LLMs on the task of automated KG-to-KG SPARQL translation. To the best of our knowledge, this is the first work to present an analysis of the cross-KG SPARQL-to-SPARQL translation task using LLMs. The code and data used in this study are publicly available and can be accessed at \url{https://github.com/semantic-systems/Automatic-SPARQL-translation}.

\section{Related Work}\label{sec:RelatedWork}
Combining multiple KGs can create richer, more comprehensive datasets by filling knowledge gaps and fostering cross-domain applications essential for tackling complex, interdisciplinary societal challenges~\citep{Caufield.2023}. However, their effectiveness is heavily depending on data reliability~\citep{Nourhan.2024} and can lead to errors or outdated information~\citep{Hofer.2024}. Additionally, heterogeneous ontologies, data formats, and languages complicate data integration~\citep{Zeng.2021}. While representation learning and graph embeddings have improved alignment accuracy by exploiting structural and semantic cues, full automation of these processes remains elusive~\citep{Savas.2023}. As KGs become more diverse, translating SPARQL queries across their heterogeneous schemas is increasingly critical~\citep{Freitas.2020}. Despite existing methods for aligning different ontologies, entity names, and predicate vocabularies to improve interoperability, this translation remains a significant challenge~\citep{Khan.2023}.

Recent works have introduced datasets, tools, and methodologies that streamline cross-KG query execution~\citep{Azevedo.2023, Kejriwal.2022}. However, dedicated frameworks for direct SPARQL-to-SPARQL translation remain scarce. Much of the existing literature focuses instead on translating SPARQL into other forms, such as converting SPARQL to SQL for querying relational databases~\citep{Chebotko.2009} or translating SPARQL to natural language for query verbalization~\citep{Ngomo.2013}. Conversely, another line of research has explored generating SPARQL queries from natural language inputs (e.g.,~\citep{Yin.2021}).
Despite these advances, systematic approaches for SPARQL-to-SPARQL translation, designed specifically to enable transparent querying across multiple KGs, remain under-explored. 

\section{Methodology}\label{sec:Methodology}
This study systematically evaluates LLM capabilities for automated SPARQL query translation between different KGs, focusing on LLM performance, methodological impacts, and translation challenges. The core approach involved: (1) constructing benchmarks for primary (DBpedia $\leftrightarrow$ Wikidata) and generalization (DBLP $\rightarrow$ OpenAlex) translation tasks; (2) systematically aligning schema elements (entities, relations) for each KG pair; (3) selecting diverse LLMs and designing varied prompting strategies (zero-shot, few-shot, Chain-of-Thought (CoT)); (4) evaluating LLM-generated translations against gold standards via exact result match; and (5) performing in-depth error classification and analysis. This multi-stage methodology allows a detailed assessment of LLM-driven SPARQL-to-SPARQL translation.

\subsection{Primary Task: DBpedia $\leftrightarrow$ Wikidata Translation}\label{ssec:PrimaryTask}

The core of the investigation centered on automated translation between DBpedia and Wikidata.

\textbf{Wikidata}~\citep{Vrandecic.2014} is a collaboratively edited, multilingual knowledge base hosted by the Wikimedia Foundation. It organizes information into items (entities, e.g., \texttt{wd:Q76} for Barack Obama) and properties (e.g., \texttt{wdt:P19} for place of birth), employing a statement-based data model that allows for rich metadata, including qualifiers, ranks, and references for individual facts, contrasting with DBpedia's typical representation.

\textbf{DBpedia}~\citep{Lehmann.2015} is a community-driven effort to extract structured information from Wikipedia, creating a large, multilingual KG that is a cornerstone of the Linked Open Data cloud. It primarily uses RDF triples and human-readable IRIs (e.g., \texttt{dbr:Barack\_Obama}, \texttt{dbo:birthPlace}). Its ontology is largely derived from Wikipedia infoboxes and categories, resulting in broad coverage but a sometimes less formally consistent structure compared to Wikidata; facts are typically represented as single, unqualified triples.

\noindent These KGs were selected due to their widespread adoption, extensive content, and, crucially, their differences in data modeling, schema organization, and entity identifier schemes (Wikidata's numeric Q/P-IDs versus DBpedia's human-readable IRIs). These distinctions present representative and substantial challenges ideal for testing automated query translation.

\noindent\textbf{QALD-9-Plus Dataset Adaptation.} The primary benchmark was derived from the QALD-9-Plus dataset~\citep{Perevalov.2022}, which provides NLQs with corresponding SPARQL queries for both DBpedia and Wikidata. For this study, English-language questions from the QALD-9-Plus training split were considered (see \autoref{tab:dataset-splits-methodology-expanded}).
\begin{table}[ht!]
\centering
\begin{threeparttable}
  \caption{Original distribution of English questions and SPARQL queries in the QALD-9-Plus dataset \\ splits, and the size of the final benchmark derived for this study.}
  \label{tab:dataset-splits-methodology-expanded}
  \begin{tabular}{lccc}
    \toprule
    \textbf{Dataset Split Source} & \textbf{English Questions} & \textbf{DBpedia Queries} & \textbf{Wikidata Queries} \\
    \midrule
    QALD-9-Plus Train & 408 & 408 & 371 \\
    QALD-9-Plus Test  & 150 & 150 & 136 \\
    \textbf{Final Benchmark (from Train)} & \textbf{100} & \textbf{100} & \textbf{100} \\
    \bottomrule
  \end{tabular}
\end{threeparttable}
\end{table}

\begin{figure}[H]
    \centering
    \includegraphics[width=\textwidth]{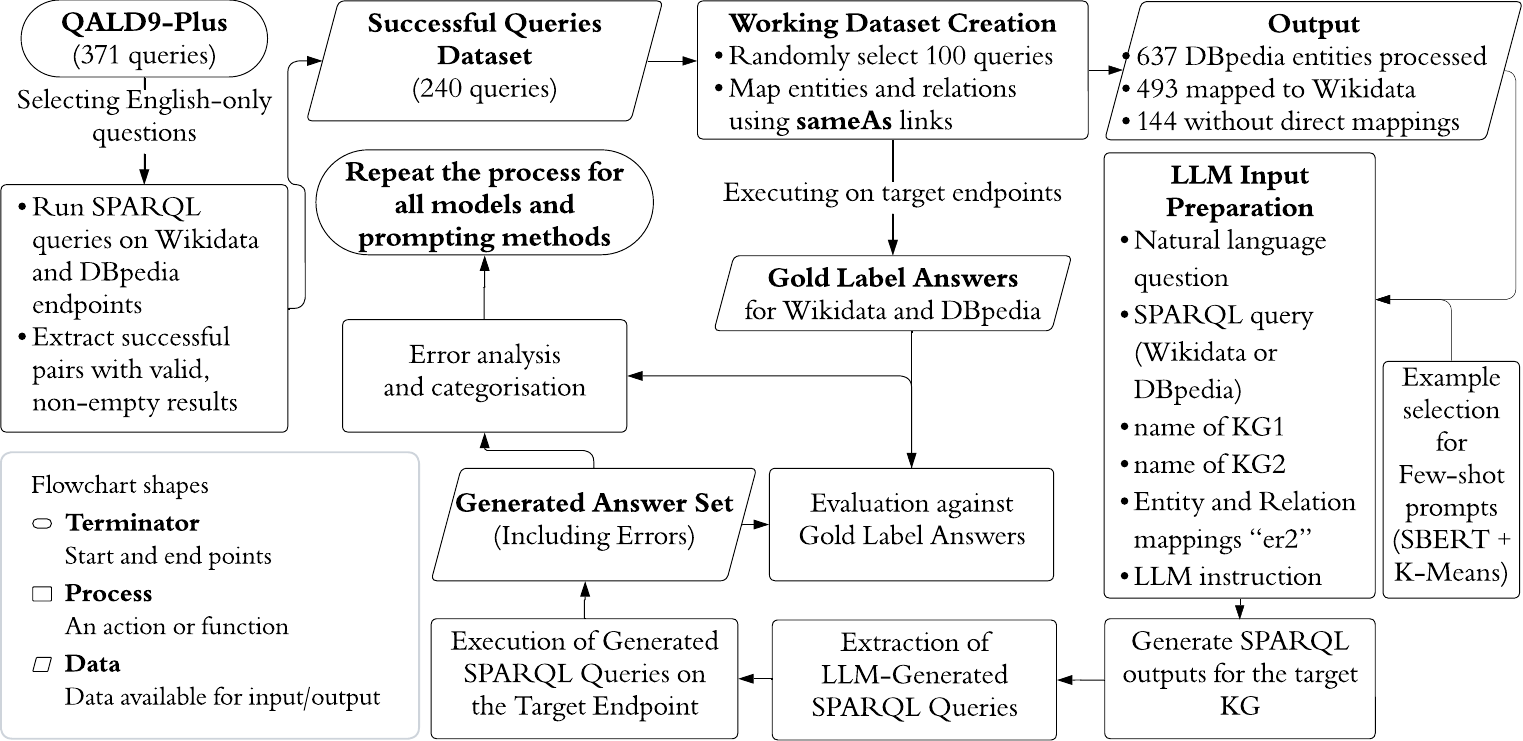}
    \caption{Workflow for Selecting, Preparing, and Translating the QALD-9-Plus derived benchmark dataset.}
    \label{fig:flow_qald9_methodology_expanded}
\end{figure}

\noindent From the QALD-9-Plus training queries that successfully executed on both Wikidata and DBpedia and returned non-empty, comparable results, a final benchmark subset of 100 NLQ-query pairs was selected. This sample size was chosen to strike a balance: it is sufficiently large to ensure a representative distribution across different query types and complexities (as detailed in Section \ref{ssec:NLQCategorization}), while still being manageable for the in-depth manual error classification and qualitative analysis necessary to understand translation errors. Gold-standard answers for these 100 queries were generated by executing the original QALD-9 queries against stable, local snapshots\footnote{The specific data dumps used for the experiments are publicly available at: \url{https://github.com/semantic-systems/Automatic-SPARQL-translation}} (reflecting data as of end-2024) of DBpedia and Wikidata, using Virtuoso \footnote{\url{https://vos.openlinksw.com/owiki/wiki/VOS/VOSSparqlProtocol}} triple stores. This ensured reproducibility by mitigating issues from evolving online data or endpoint instability. The overall workflow is illustrated in \autoref{fig:flow_qald9_methodology_expanded}.

\noindent To ensure transparency and facilitate further research, the curated benchmark datasets, alignment scripts, query sets, and evaluation code used in this study are publicly available\footnote{\url{https://github.com/semantic-systems/Automatic-SPARQL-translation}}.

\subsection{Generalization Task: DBLP $\rightarrow$ OpenAlex Translation}\label{ssec:GeneralizationTask}
This task assesses the generalizability of translation methods for a domain-specific KG pair in the scientific communication domain, using KGs with distinct modeling characteristics from DBpedia and Wikidata.

\textbf{DBLP}~\citep{Ley.2002} is a highly curated bibliographic KG for computer science, indexing millions of publications and authors. It features a uniform semantic data model and importantly utilizes ORCID identifiers for author disambiguation, which simplifies linking between graphs.

\textbf{OpenAlex}~\citep{Priem.2022} is a large, fully open scholarly KG aggregating metadata on academic works, authors, institutions, and concepts. It also integrates ORCID identifiers and links to Wikidata concepts, aiming for broad coverage and interoperability.

\textbf{DBLP-QuAD Dataset Adaptation.} A 100-query benchmark was created by adapting data from the DBLP-QuAD dataset~\citep{Banerjee.2023}. This involved selecting 100 NLQ-SPARQL query pairs from DBLP-QuAD based on query templates semantically translatable to OpenAlex (e.g., excluding queries for BibTeX types or DBLP-specific metadata not present in OpenAlex). For these, semantically equivalent OpenAlex SPARQL queries were manually created, relying heavily on ORCID identifiers for accurate author mapping and on the careful alignment of DBLP relations (e.g., \texttt{dblp:authoredBy}) to their OpenAlex counterparts (e.g., \texttt{oa:hasAuthorship} linked to \texttt{oa:hasAuthor}). Gold-standard answers were obtained from the respective official SPARQL endpoints\footnote{\url{https://semopenalex.org/sparql}}\footnote{\url{https://sparql.dblp.org/}} during February 2025, ensuring the results reflect the state of the KGs at that specific time.

\subsection{Entity and Relationship Mapping}\label{ssec:ERMappingExpanded}
Explicit entity-relationship mapping was considered a critical factor for translation accuracy. Alignment data was prepared for relevant prompting strategies.

\textbf{DBpedia $\leftrightarrow$ Wikidata Alignment.} A systematic, multi-step process generated these alignments: (a) DBpedia query prefixes (e.g., \texttt{dbo:}) in benchmark queries were expanded to full URIs. (b) Unique DBpedia entity and relation URIs were extracted using regular expressions. (c) These DBpedia URIs were mapped to Wikidata equivalents by querying DBpedia for explicit \texttt{owl:sameAs} (for entities), \texttt{owl:equivalentPro-\\perty} (for relations), and \texttt{owl:equivalentClass} links. Only valid Wikidata URI mappings were retained. From 637 unique DBpedia terms in the benchmark, 493 were successfully mapped; the 144 unmapped (often due to structural differences or lack of direct equivalences) were reviewed, and queries with them retained to test LLM robustness in such cases.

\textbf{DBLP $\rightarrow$ OpenAlex Mapping} relied on shared ORCIDs for authors and manual relation mapping during gold-standard OpenAlex query creation.
Resulting mappings were structured (typically JSON mapping source IRIs to target IRIs) and provided to LLMs as the entity-relation mapping variable \texttt{er2}. An example \texttt{er2} structure is: \\ \texttt{\{"dbpedia\_id": "\url{http://dbpedia.org/ontology/director}", \\ "wikidata\_id": ["\url{http://www.wikidata.org/entity/P57}"]\}}.

\subsection{Evaluation Framework}\label{ssec:EvaluationFramework}
\noindent\textbf{Correctness Evaluation.} LLM-generated SPARQL translations were deemed correct if their executed result sets precisely matched pre-established gold-standard answers (disregarding order unless inherently meaningful for ranked and ordered queries). This strict exact match criterion provides an objective measure of functional equivalence.

\noindent\textbf{Error Analysis.} To investigate translation failures, an 8-category error framework was adapted and further extended from previous work~\citep{Azmy.2018, Banerjee.2022}. The categories, detailed below, capture common structural and semantic issues:
\begin{itemize}
    \setlength\itemsep{0.05em}
    \item \textbf{Unadapted Dataset Patterns:} The translated query incorrectly reuses IRIs, properties, or schema prefixes from the source KG instead of those appropriate for the target KG.
    \item \textbf{Query Bad Formed Error:} The query fails SPARQL syntax parsing entirely, rendering it inexecutable by the target endpoint.
    \item \textbf{Ontology Treated as Resource / Property Treated as Entity:} A DBpedia ontology class or a Wikidata property is mistakenly used in a position where a DBpedia resource or a Wikidata entity (item) is expected.
    \item \textbf{Resource Treated as Ontology / Entity Treated as Property:} Conversely, a DBpedia resource or a Wikidata entity (item) is incorrectly used where a DBpedia ontology class/property or a Wikidata property is expected.
    \item \textbf{Wikidata Missing \texttt{P31} / DBpedia Missing \texttt{rdf:type}:} Essential class typing information is omitted; for instance, Wikidata's crucial "instance of" (\texttt{wdt:P31}) property or DBpedia's standard \texttt{rdf:type} for class membership is missing when required.
    \item \textbf{Wrong or Missing Ontology / Wrong or Missing Property:} The query employs incorrect DBpedia ontology classes or Wikidata properties, or omits essential ones (or includes superfluous ones) needed to correctly fulfill the query's intent.
    \item \textbf{Wrong or Missing Resource / Wrong or Missing Entity:} The query incorrectly specifies or omits necessary DBpedia resources or Wikidata entities (items), leading to semantically flawed results.
    \item \textbf{Structural Error:} The query is syntactically valid but its logical structure (e.g., triple patterns, filter logic, or prefix declarations not covered by unadapted patterns) is inconsistent with the target KG's actual data model or schema constraints, typically yielding empty or unintended results.
\end{itemize}
\noindent Each incorrect query could receive multiple error labels, reflecting combined failure modes. Classification was hybrid: automated pre-screening using heuristics,  followed by manual review and judgement by the researchers to ensure high reliability.

\subsection{Natural Language Question Categorization}\label{ssec:NLQCategorization}
To enable a more nuanced error analysis based on query intent, the 100 primary benchmark NLQs were manually categorized by their linguistic structure and expected answer type (see Table~\ref{tab:question_categories_dist}). These 100 questions were randomly selected from the successfully executing queries within the QALD-9-Plus training set and were chosen to ensure a representative distribution across the different query types and complexities in the dataset. This categorization allowed for the correlation of error patterns with question complexity.
\begin{table}[H]
\centering
\caption{Categorization and Distribution of Natural Language Questions with Examples.}
\label{tab:question_categories_dist}
\begin{tabular}{lcp{8cm}}
\toprule
\textbf{Category} & \textbf{Count} & \textbf{Examples} \\
\midrule
Single Fact & 34 & "When was Barack Obama born?" \newline "Where is the headquarters of Google?" \\
Comprehensive List & 18 & "List all countries in South America." \newline "Which cities have hosted the Olympic Games?" \\
Aggregated List & 14 & "Which books were written by Agatha Christie?" \newline "Which people were born in Berlin?" \\
Single Person & 14 & "Who is the president of France?" \newline "Who discovered penicillin?" \\
Rank or Ordered Info. & 10 & "What is the tallest mountain in the world?" \newline "Who are the top five richest people?" \\
Numerical Count & 6 & "How many children did Albert Einstein have?" \newline "What is the population of Germany?" \\
Filtered Multi-Entity & 4 & "Which cities hosted both Summer and Winter Olympics?" \newline "Which actors worked with both Tarantino and Scorsese?" \\
\bottomrule
\end{tabular}
\end{table}

\section{Experimental Setup}\label{sec:ExperimentalSetup}
This section details the specific LLMs, the design and application of prompting strategies, and the procedures for output processing used to conduct the SPARQL query translation experiments. 

\subsection{Large Language Models Evaluated}\label{ssec:LLMsEvaluated}
Three distinct, openly accessible LLMs were selected to investigate the influence of model scale, architecture, and reasoning capabilities on SPARQL translation accuracy. These models represent a spectrum of parameter sizes and reported strengths, chosen for their proven performance in NLP benchmarks and suitability for structured query tasks:

First, \textbf{Llama 3.1-8B Instruct}~\citep{Dubey.2024}, developed by Meta, served as a representative of high-performing smaller models. With 8 billion parameters and an extended context window of up to 128,000 tokens, it is specifically fine-tuned for instruction-following and has demonstrated proficiency in structured generation tasks such as coding and formal query formulation. Its inclusion allows for an assessment of how well more compact, yet capable and reasoning-aware, models handle the complexities of cross-schema translation.

Second, \textbf{Mistral-Large-Instruct-2407}~\citep{Mistral.2024} from Mistral AI was selected due to its substantial scale (123 billion parameters) and an extensive context window of 128,000 tokens. This model's capacity for context comprehension and nuanced reasoning, potentially supported by mechanisms like Grouped-Query Attention for efficiency, was deemed particularly beneficial for translating detailed SPARQL queries involving complex textual contexts from NLQs and intricate entity-relationship mappings. It represents the upper end of openly accessible model sizes used in this study.

Third, \textbf{DeepSeek-R1-Distill-Llama-70B}~\citep{Guo.2025}, a 70-billion parameter model distilled from the Llama-3.3-70B-Instruct architecture, was incorporated into the experiments. Positioned between Llama 3.1-8B and Mistral-Large-Instruct-2407 in terms of parameter count, DeepSeek-R1-Distill-Llama-70B is recognized for state-of-the-art performance across numerous NLP benchmarks and, importantly for this research, its specialized design for advanced logical reasoning and structured data modeling, making it particularly well-suited for CoT prompting evaluations.

\subsection{Prompting Strategies and Translation Procedures}\label{ssec:PromptingStrategies}
A series of prompting strategies were designed and systematically applied to guide the selected LLMs in the SPARQL query translation tasks. These strategies ranged from minimal guidance (zero-shot) to more structured approaches involving in-context examples (few-shot) and explicit intermediate reasoning steps (CoT), allowing for a thorough investigation of how prompt engineering affects translation performance.

\subsubsection{Core Prompt Design}\label{sssec:CorePromptDesign}
A consistent core structure, adapted for each prompting method, was maintained for all prompts. Each prompt provided to the LLM included: (a) the NLQ to be translated; (b) the complete source SPARQL query from the initial knowledge graph (KG1), acting as a reference; (c) the names of both the source KG (KG1) and the target KG (KG2) (e.g., "DBpedia" and "Wikidata") to contextualize the task; and (d) for strategies using explicit schema information, a structured representation of entity and relationship mappings between KG1 and KG2 (referred to as `er2`). A critical instruction common to all prompts was to request the LLM to enclose the final, complete translated SPARQL query for KG2 within \texttt{<sparql>} and \texttt{</sparql>} tags, a measure implemented to facilitate robust automated extraction of the query from potentially verbose LLM outputs. \\
\noindent Example prompt for few-shot translation from DBpedia to Wikidata: \\ 
\phantomsection\label{ex:PromptFewShot}
\vspace{-0.5\baselineskip}
\noindent\fbox{%
    \parbox{\dimexpr\linewidth-3\fboxsep-1\fboxrule\relax}{
        \vspace{\fboxsep}
        \texttt{\footnotesize%
        \{"\textbf{natural\_language\_question}": "Which films did Stanley Kubrick direct?", \\
        "\textbf{sparql\_query\_kg1}": "PREFIX dbo: \textless http://dbpedia.org/ontology/\textgreater{} PREFIX res: \textless http://dbpedia.org/resource/\textgreater{} SELECT DISTINCT ?uri WHERE \{ ?uri dbo:director res:Stanley\_Kubrick \}", \\
        "\textbf{kg1\_name}": "DBpedia", "\textbf{kg2\_name}": "Wikidata", \\
        "\textbf{er2}": [\{"\textbf{dbpedia\_id}": "http://dbpedia.org/ontology/director", \\ "\textbf{wikidata\_ids}": ["http://www.wikidata.org/entity/P57"]\}, \\
        \{"\textbf{dbpedia\_id}": "http://dbpedia.org/resource/Stanley\_Kubrick", \\ "\textbf{wikidata\_ids}": ["http://www.wikidata.org/entity/Q2001"]\}], \\
        "\textbf{instruction}": "Given the information above, produce a SPARQL query for KG2. In your answer please highlight the final, complete SPARQL query within the tags '\textless sparql\textgreater ' and '\textless /sparql\textgreater '. Here are 4 examples:" \\ \textit{ (For few-shot prompting, four translation examples would follow here.)}}
        }%
        \vspace{\fboxsep}
    }%
\subsubsection{Prompting for DBpedia $\leftrightarrow$ Wikidata Translation}\label{sssec:PromptingDBpediaWikidata}
For the primary translation task between DBpedia and Wikidata, five distinct prompting methods were evaluated:

\textbf{Zero-Shot Prompting (Baseline):} This approach established a baseline by assessing the LLMs' inherent ability to translate SPARQL queries without any task-specific examples and, crucially, without the explicit entity-relation (ER) mapping. The prompt contained only the NLQ, source query, KG names, and output instruction. This setup, applied to Llama 3.1-8B and Mistral-Large-Instruct-2407, was expected to highlight challenges LLMs face when relying solely on pre-trained knowledge.

\textbf{Zero-Shot Prompting with Entity-Relation Mapping:} To mitigate baseline limitations, particularly ambiguity in mapping schema elements, this variant augmented the zero-shot prompt by including the ER mapping variable. This variable provided an explicit, JSON-formatted mapping of corresponding entities/relations between DBpedia and Wikidata (generated as detailed in Section \ref{ssec:ERMappingExpanded}), aiming to directly quantify the impact of schema alignment information when applied to Llama 3.1-8B and Mistral-Large-Instruct-2407.

\textbf{Few-Shot Prompting:} This strategy aimed to enhance accuracy by providing four complete, illustrative DBpedia-Wikidata translation examples within the prompt. Each example comprised an NLQ, its KG1/KG2 SPARQL query examples, and the relevant ER mapping. These examples were carefully selected from remaining non-test QALD-9-Plus data (ensuring no overlap with the 100 test queries to prevent data leakage) using Sentence-BERT (SBERT)~\citep{Reimers.2019} embeddings and K-Means clustering to ensure diversity across query types. The prompt also included its specific ER mapping, again applied to both models Llama 3.1-8B and Mistral-Large-Instruct-2407.

\textbf{Chain-of-Thought (CoT) Prompting:} To explore the impact of explicit reasoning, CoT prompting~\citep{Wei.2022} instructed LLMs to first articulate a step-by-step explanation of their reasoning for constructing the target query (detailing query part formation and entity/property choices based on source query and ER mappings) before providing the final query. This method, aimed at encouraging deliberate planning, was tested across all three LLMs.

\textbf{Chain-of-Thought Prompting with Tags:} This structured CoT variant explicitly guided LLMs through a predefined sequence of five cognitive sub-tasks using demarcated \texttt{<think>...</think>} tags: (a) identify key entities/relations in the NLQ and map them using `er2`; (b) analyze the source SPARQL query structure; (c) find equivalent target KG properties using mappings; (d) construct the target SPARQL query maintaining logical structure; and (e) conceptually validate the query against the target KG's model. This approach, also including ER mapping, aimed for enhanced interpretability and consistency applied to all three LLMs.

\subsubsection{Experimental Procedure for DBLP $\rightarrow$ OpenAlex Translation}\label{sssec:SetupDblpOpenalex}
To assess generalizability to the specialized scholarly domain, and given DBLP/OpenAlex's potentially lower prevalence in LLM training data, two prompting strategies incorporating explicit schema guidance (ER) were applied to Llama 3.1-8B and Mistral-Large-Instruct-2407:

\textbf{Zero-Shot Prompting with Entity-Relation Mapping (for DBLP-OpenAlex):} This approach directly incorporated ER mappings (ORCID links and manually defined DBLP-to-OpenAlex relations) from the outset, deemed essential for a fair baseline given the KGs' specificity.

\textbf{Few-Shot Prompting (for DBLP-OpenAlex):} This provided four curated DBLP-to-OpenAlex translation examples (selected via SBERT/K-Means from non-test adapted DBLP-QuAD queries, covering diverse scholarly patterns like temporal filters, co-authorship, and multi-variable queries), alongside the ER mapping for the test query, evaluating in-context learning for domain generalization.

\subsection{Result Extraction and Post-processing}\label{ssec:ResultExtractionExp}
To reliably isolate executable SPARQL queries from raw LLM outputs, which often interleave queries with ancillary text or formatting artifacts, a resilient post-processing pipeline was implemented. This pipeline primarily searched for content within the instructed \texttt{<sparql>} tags, using fallbacks (e.g., detecting markdown code blocks, identifying SPARQL keyword-initiated segments) if necessary. Candidate queries then underwent automated validation (heuristic checks for essential clauses like \texttt{SELECT} and \texttt{WHERE}) and thorough cleaning (e.g., removing extraneous tags, normalizing whitespace). Queries failing automated extraction or validation were logged. A two-stage manual verification then reviewed logged failures and subsequently checked all successfully processed queries for structural integrity before execution.

\section{Results}\label{sec:Results}
This section presents empirical findings on SPARQL query translation accuracy and error patterns for the primary DBpedia~$\leftrightarrow$~Wikidata task and the DBLP~$\rightarrow$~OpenAlex generalization task.

\begin{figure}[htbp]
    \centering
    \includegraphics[width=1\linewidth]{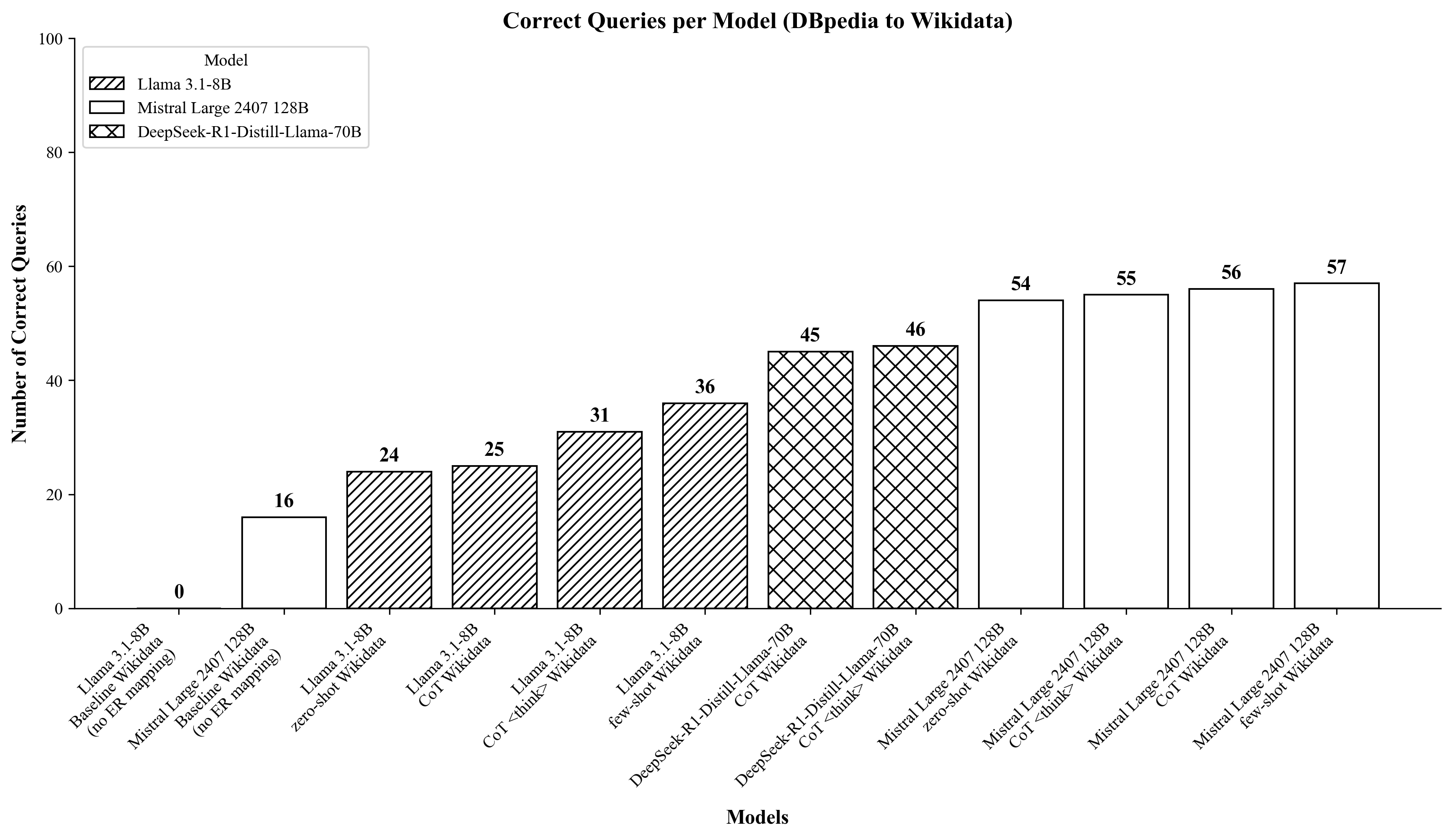}
    \caption{Correctly Translated Queries: Model \& Strategy (DBpedia$\rightarrow$Wikidata; N=100)}
    \label{fig:correct_queries_wikidata}
\end{figure}

\begin{figure}[htbp]
    \centering
    \includegraphics[width=1\linewidth]{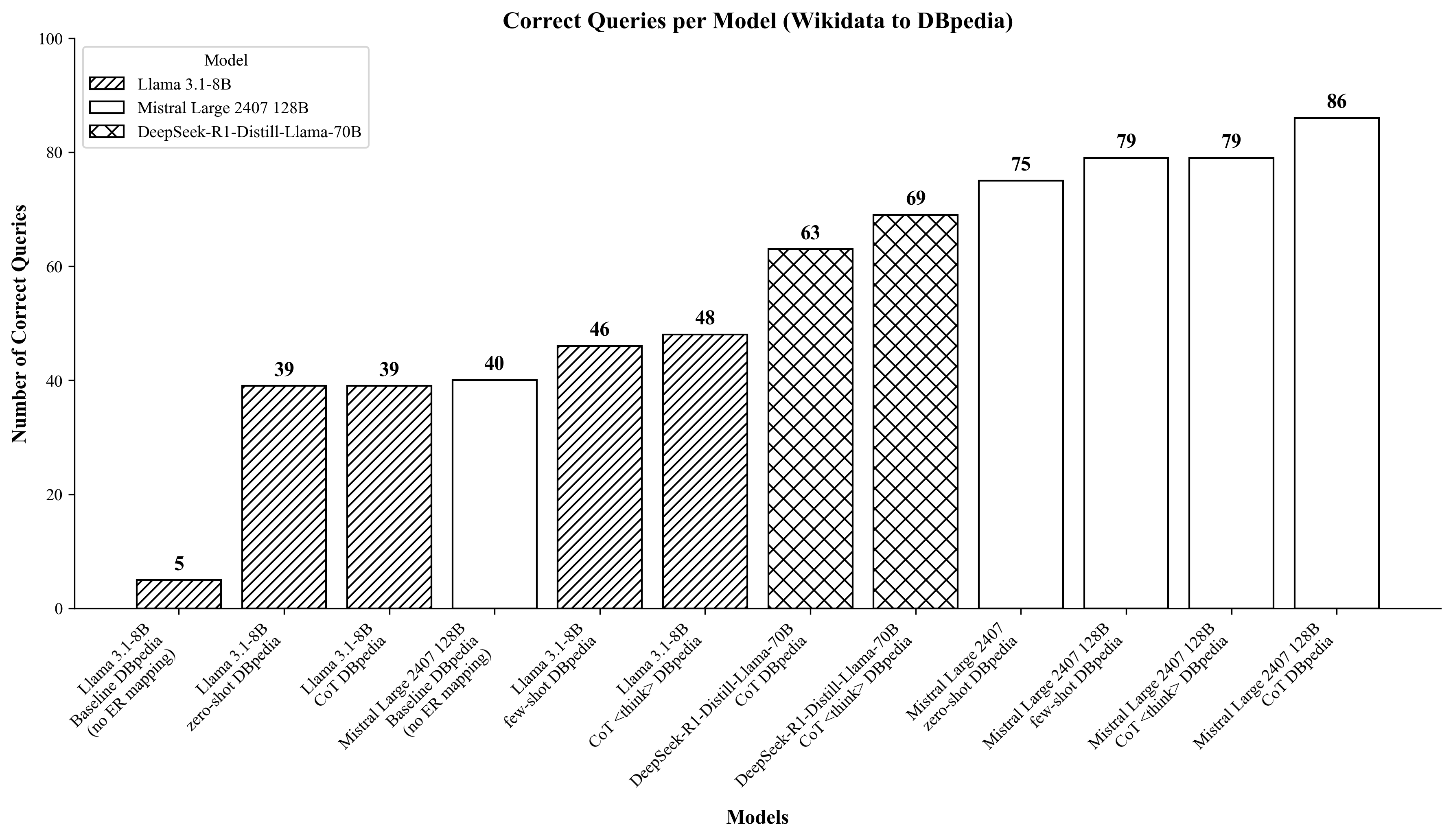}
    \caption{Correctly Translated Queries: Model \& Strategy (Wikidata$\rightarrow$DBpedia; N=100)}
    \label{fig:correct_queries_dbpedia}
\end{figure}

\subsection{Translation Accuracy: DBpedia $\leftrightarrow$ Wikidata}\label{ssec:TranslationAccuracy}

\noindent\textbf{DBpedia to Wikidata Translation.}
Accuracy for DBpedia~$\rightarrow$~Wikidata translations varied significantly (\autoref{fig:correct_queries_wikidata}). The Llama 3.1-8B baseline (zero-shot without ER mapping) achieved 0\% accuracy. Performance improved with ER mapping and structured prompting, with Llama 3.1-8B reaching 36\% (few-shot). DeepSeek-R1-Distill-Llama-70B achieved up to 46\% (CoT with \texttt{<think>} tags). Mistral-Large-Instruct-2407 was the strongest, peaking at 57\% (few-shot with ER mapping), a substantial improvement over its 16\% baseline. Structured prompting with ER mapping consistently outperformed simpler approaches.

\noindent\textbf{Wikidata to DBpedia Translation.}
Translations from Wikidata~$\rightarrow$~DBpedia generally yielded higher accuracies (\autoref{fig:correct_queries_dbpedia}). Llama 3.1-8B improved from a 5\% baseline to 48\% (CoT \texttt{<think>} with ER mapping). DeepSeek-R1-Distill-Llama-70B reached up to 69\% (CoT \texttt{<think>} tags). Mistral-Large-Instruct-2407 excelled, achieving 86\% accuracy with its few-shot prompting approach (including ER mapping); its zero-shot variant with ER mapping (75\%) also surpassed all Llama 3.1-8B configurations.

\subsection{Overview of Translation Errors (DBpedia $\leftrightarrow$ Wikidata)}\label{ssec:TranslationErrors}
For the DBpedia-Wikidata tasks, each of the 100 benchmark questions was processed using 12 distinct combinations of LLMs and prompting strategies for both translation directions (DBpedia$\leftrightarrow$Wikidata). This resulted in a total of 2400 model-query runs, across which 1629 error instances were logged for translations targeting Wikidata and 1029 for those targeting DBpedia. Notably, many queries exhibited multiple error types simultaneously. \autoref{tab:all_errors_db_wd_condensed} details the distribution of the eight defined error categories. \texttt{Structural Error} was the most prevalent category in both translation directions.

\begin{table}[ht!]
\centering
\caption{Distribution of Error Types in DBpedia $\leftrightarrow$ Wikidata Translations. Counts represent total instances per category, aggregated from N=100 unique base queries tested across 12 distinct model/prompt configurations for each translation direction.}
\label{tab:all_errors_db_wd_condensed}
\begin{tabular}{p{0.49\linewidth}rr}
\toprule
\textbf{Error Category} & \textbf{Target KG: Wikidata} & \textbf{Target KG: DBpedia} \\
\midrule
\texttt{Structural Error} & 534 & 483 \\
\texttt{Wrong Entity / Resource} & 351 & 71 \\
\texttt{Wrong Property / Ontology} & 287 & 148 \\
\texttt{Query Bad Formed Error} & 261 & 201 \\
\texttt{Missing P31 / Missing rdf:type} & 89 & 47 \\
\texttt{Unadapted Dataset Patterns} & 44 & 34 \\
\texttt{Property Treated as Entity / Ontology Treated as Resource} & 47 & 8 \\
\texttt{Entity Treated as Property / Resource Treated as Ontology} & 16 & 37 \\
\midrule
\textbf{Total Error Instances Logged} & \textbf{1629} & \textbf{1029} \\
\bottomrule
\end{tabular}
\end{table}

\noindent A key finding from the detailed error analysis was the strong co-occurrence of certain error types indicating that errors rarely appeared in isolation. For example, \texttt{Missing P31} errors (to Wikidata) and \texttt{Missing rdf:type} errors (to DBpedia) were almost always (97-98\% of instances) accompanied by a \texttt{Structural Error}. This pattern extended to other categories; \texttt{Query Bad Formed} errors also frequently co-occurred with \texttt{Structural Error} (e.g., 72.8\% for Wikidata target, 97.5\% for DBpedia target), and incorrect entity mappings (\texttt{Wrong Entity/Resource}) also showed a strong association with structural issues (e.g., 78.4\% for Wikidata target). This suggests that initial schema mapping mistakes often cascade, leading to broader structural inconsistencies and often resulting in queries exhibiting multiple distinct error types, indeed, a majority of incorrect queries were assigned two or more error labels. Furthermore, the complexity of the NLQ influenced error rates. Simpler question types (e.g., \textit{Single Fact}, \textit{Numerical Count}) averaged fewer errors per query. Conversely, complex types requiring aggregation, filtering, or ordering consistently exhibited a higher average number of distinct errors; for instance, \textit{Comprehensive List} questions averaged the most errors when translating to Wikidata, while \textit{Ordered Information} questions were most problematic for DBpedia translations. These interconnected patterns underscore persistent challenges in accurate logical and semantic mapping, especially for complex intents and in scenarios where initial semantic misalignments can corrupt the entire query structure.

\subsection{Generalization Accuracy: DBLP $\rightarrow$ OpenAlex}\label{ssec:GeneralizationAccuracy}
Experiments translating DBLP queries to OpenAlex highlighted the impact of prompting strategy on generalizability to a specialized domain. As shown in \autoref{tab:dblp_openalex_accuracy}, with \textbf{zero-shot prompting} (including ER mapping), performance was poor: Llama 3.1-8B achieved only 1\% accuracy, and Mistral-Large-Instruct-2407 6\%. Many queries failed execution or yielded no answer. In contrast, \textbf{few-shot prompting} with ER mapping significantly boosted performance, with Llama 3.1-8B reaching 43\% and Mistral-Large-Instruct-2407 achieving 86\% accuracy. This underscores the critical role of few-shot examples and ER mapping for effective translation to less common or specialized KG schemas.

\begin{table}[ht!]
\centering
\caption{Generalization Accuracy: DBLP $\rightarrow$ OpenAlex Translation Results (N=100 queries per configuration).}
\label{tab:dblp_openalex_accuracy}
\begin{tabular}{llrr}
\toprule
\textbf{Model Name} & \textbf{Prompting Strategy} & \textbf{Correct} & \textbf{Incorrect / Failed} \\
\midrule
Llama 3.1-8B Instruct & Zero-shot (with ER mapping) & 1 & 99 \\
Llama 3.1-8B Instruct & Few-shot (with ER mapping)  & 43 & 57 \\
\midrule
Mistral-Large-Instruct-2407 & Zero-shot (with ER mapping) & 6 & 94 \\
Mistral-Large-Instruct-2407 & Few-shot (with ER mapping)  & 86 & 14 \\
\bottomrule
\end{tabular}
\end{table}

\section{Discussion}\label{sec:Discussion}
The results demonstrate that contemporary LLMs, when appropriately guided, can achieve high accuracy in translating SPARQL queries between heterogeneous KGs; however, performance is significantly influenced by model capacity, prompting strategy, and the provision of schema alignments.

\textbf{Impact of Model Size and Architecture}:
The findings show a correlation between model size and performance. The larger Mistral-Large-Instruct-2407 (123B parameters) consistently outperformed DeepSeek-R1-Distill-Llama-70B, which in turn surpassed Llama 3.1-8B. This indicates that increased model scale provides greater representational capacity crucial for understanding complex query structures and nuanced semantic mappings between disparate KG schemas, as evidenced by Mistral-Large's stronger baseline performance compared to Llama 3.1-8B's effort, even with ER mapping.

\textbf{Effectiveness of Prompting Strategies and Schema Mapping}:
Structured prompting methods, specifically few-shot and CoT prompting, consistently surpassed zero-shot approaches, even those with ER mappings, often by wide margins.
The critical role of providing explicit entity and relationship mappings was also clearly demonstrated. Accuracy collapsed in baseline runs without explicit mappings (e.g., Llama 3.1-8B: 0\% for DBpedia$\rightarrow$Wikidata). Supplying ER mapping tables boosted accuracy by over fifty percentage points in many cases, enabling models to focus on structural transformation rather than guessing identifiers.

\textbf{Performance of DeepSeek-R1-Distill-Llama-70B with CoT}:
The study also examined the specific CoT performance of DeepSeek-R1-Distill-Llama-70B, as this model is specifically recognized for its advanced logical reasoning and structured data modeling capabilities. While the model outperformed Llama 3.1-8B in CoT tasks, it was consistently surpassed by the larger Mistral-Large-Instruct-2407 model. The explicit \texttt{<think>} tags did not yield a notable additional boost for DeepSeek-R1-Distill-Llama-70B, suggesting its inherent reasoning is well-leveraged by standard CoT, or that model scale remains a more dominant factor than specific CoT enhancements for this translation task.

\textbf{Interpreting Translation Asymmetries and Error Patterns:}
Translations from Wikidata to DBpedia were consistently more accurate. This asymmetry likely stems from DBpedia's human-readable IRIs (aligning better with NLQs) and potentially greater pre-training exposure, compared to Wikidata's abstract numeric identifiers.

The predominance of \texttt{Structural Error} (30-50\% of instances, see \autoref{tab:all_errors_db_wd_condensed}) is significant. These syntactically valid but logically flawed queries often co-occurred with semantic issues like \texttt{Wrong Entity/Property} or missing type definitions (e.g., \texttt{P31} or \texttt{rdf:type}), with co-occurrence rates for missing types being high (97-98\%). This suggests initial mapping misalignments frequently cascade, corrupting entire query structures. Most incorrect queries indeed exhibited multiple error types. 
Finally, simpler NLQ types (e.g., single fact, numerical count) averaged fewer errors than complex queries requiring aggregation, filtering, or ordering.

\textbf{Generalization and Implications for Interoperability}:
The DBLP $\rightarrow$ OpenAlex experiments (1-6\% zero-shot vs. 86\% few-shot accuracy for Mistral-Large-Instruct-2407) demonstrated that while zero-shot translation struggles in specialized domains, few-shot prompting with ER mapping dramatically improves performance. This strong result indicates a high potential for the approach to generalize to other structured, domain-specific KGs.

\textbf{Broader Implications and Practical Recommendations:}
This study demonstrates that LLMs offer a viable pathway for automating SPARQL query translation, which can substantially reduce the manual effort for organizations managing multiple KGs. The key "practical recipe" emerging from this research for achieving effective translation is to: 
\begin{itemize}
    \setlength\itemsep{0.0em}
    \item Using large-capacity LLMs for the translation task.
    \item Employing structured prompting techniques, with few-shot learning (using representative examples) proving particularly effective.
    \item Ensuring models have access to accurate and up-to-date entity and relation mapping tables, as this is crucial for optimal performance.
    \item When designing new KGs or evolving existing ones, prioritizing human-language friendly identifiers (similar to those in DBpedia). This approach can simplify entity-relation mapping and improve LLM translation accuracy, as suggested by the observed translation asymmetries with numerically-identified KGs like Wikidata.
\end{itemize}
\noindent Adopting this approach can lower the barrier to KG integration, fostering broader adoption of linked data principles and enabling more extensive cross-domain knowledge discovery. While current methods advance automation, the developed error classification framework also provides a valuable tool for diagnosing remaining issues and guiding future refinements toward even more robust systems.

\section{Limitations}
The study's scope has limitations. The evaluation benchmarks, while carefully curated, were of moderate size and exclusively English-based. Furthermore, while the generalization task provides initial evidence of the method's potential beyond encyclopaedic KGs, testing on a single additional domain is not sufficient to prove the approach will generalize to any other KG. A further consideration is that the LLMs may have been exposed to the QALD-9-Plus dataset during pre-training; however, as is evident from the poor performance on the zero-shot tasks without entity alignment (e.g., 0\% accuracy for Llama 3.1-8B translating from DBpedia to Wikidata), the models do not appear to have perfect recall in such scenarios. Additionally, static KG snapshots were employed for the experiments, which do not reflect real-world KG evolution, thus limiting long-term robustness insights. Finally, the inherent stochasticity of LLM outputs means that repeated queries under identical conditions might still yield slightly different translations.

\section{Future Work}\label{sec:ConclusionFutureWork}
Building on our findings, promising future research directions include:
investigating model scaling and efficiency (balancing larger LLMs with fine-tuned smaller models, considering CO\textsubscript{2} costs);
advanced prompt engineering, such as exploring sophisticated CoT or adaptive techniques to enhance reasoning and reduce errors;
deeper analysis of translation asymmetries (particularly with numerically encoded KGs like Wikidata) and performance on complex query structures (e.g., deep nesting, aggregation);
broader generalization assessments across diverse KG domains and evaluation of robustness against KG schema evolution and data drift;
and enhancing LLM output consistency and parsability via specialized instruction or format-aware fine-tuning to reduce post-processing reliance.
Pursuing these avenues can further advance the practical application of LLMs for robust, KG-agnostic query translation, ultimately fostering greater data interoperability across the Semantic Web.

\newpage

\bibliographystyle{vancouver}
\bibliography{main}

\begin{thebibliography}{10}

\bibitem{Vrandecic.2014}
Vrande{\v{c}}i{\'c} D, Kr{\"o}tzsch M.
\newblock Wikidata.
\newblock Communications of the ACM. 2014;57(10):78-85.
\newblock doi:10.1145/2629489.

\bibitem{Lehmann.2015}
Lehmann J, Isele R, Jakob M, Jentzsch A, Kontokostas D, Mendes PN, et~al.
\newblock DBpedia -- A large-scale, multilingual knowledge base extracted from
  Wikipedia.
\newblock Semantic Web. 2015;6(2):167-95.
\newblock doi:10.3233/SW-140134.

\bibitem{Zeng.2021}
Zeng K, Li C, Hou L, Li J, Feng L.
\newblock A comprehensive survey of entity alignment for knowledge graphs.
\newblock AI Open. 2021;2:1-13.
\newblock doi:10.1016/j.aiopen.2021.02.002.

\bibitem{Schreiber.2014}
Schreiber G, Raimond Y, Manola F, Miller E, McBride B. RDF 1.1 Primer [W3C
  Recommendation]; 2014.
\newblock Accessed: February 5, 2025.
\newblock Available from: \url{https://www.w3.org/TR/rdf11-primer/}.

\bibitem{Jorge.2009}
P{\'e}rez J, Arenas M, Gutierrez C.
\newblock Semantics and complexity of SPARQL.
\newblock ACM Trans Database Syst. 2009;34(3):16:1-16:45.
\newblock doi:10.1145/1567274.1567278.

\bibitem{Ali.2022}
Ali W, Saleem M, Yao B, Hogan A, Ngomo ACN.
\newblock A survey of RDF stores {\&} SPARQL engines for querying knowledge
  graphs.
\newblock The VLDB Journal. 2022;31(3):1-26.
\newblock doi:10.1007/s0 0778-021-00711-3.

\bibitem{Hogan.2022}
Hogan A, Cochez M, de~Melo G.
\newblock Knowledge graphs. vol.~22 of Synthesis lectures on data, semantics
  and knowledge.
\newblock Cham: Springer; 2022.

\bibitem{Hofer.2024}
Hofer M, Obraczka D, Saeedi A, K{\"o}pcke H, Rahm E.
\newblock Construction of Knowledge Graphs: Current State and Challenges.
\newblock Information. 2024;15(8):509.
\newblock doi:10.3390/info15080509.

\bibitem{Ashish.2017}
Vaswani A, Shazeer N, Parmar N, Uszkoreit J, Jones L, Gomez AN, et~al.
\newblock Attention is All you Need.
\newblock In: Proceedings of the 31st International Conference on Neural
  Information Processing Systems 2017, December 4-9, 2017, Long Beach, CA, USA;
  2017. p. 6000-10.
\newblock Available from:
  \url{http://papers.nips.cc/paper/7181-attention-is-all-you-need}.

\bibitem{Brown.2020}
Brown TB, Mann B, Ryder N, Subbiah M, Kaplan J, Dhariwal P, et~al.
\newblock Language Models are Few-Shot Learners.
\newblock In: Advances in Neural Information Processing Systems 33: Annual
  Conference on Neural Information Processing Systems 2020, NeurIPS 2020,
  December 6-12; 2020. p. 1877-901.

\bibitem{Yin.2021}
Yin X, Gromann D, Rudolph S.
\newblock Neural machine translating from natural language to SPARQL.
\newblock Future Gener Comput Syst. 2021;117:510-9.
\newblock doi:10.1016/j.future.2020.12.013.

\bibitem{Agrawal.2024}
Agrawal G, Kumarage T, Alghamdi Z, Liu H.
\newblock Can Knowledge Graphs Reduce Hallucinations in LLMs? : A Survey.
\newblock In: Duh K, Gomez H, Bethard S, editors. Proceedings of the 2024
  Conference of the North American Chapter of the Association for Computational
  Linguistics: Human Language Technologies (Volume 1: Long Papers).
  Stroudsburg, PA, USA: {Association for Computational Linguistics}; 2024. p.
  3947-60.
\newblock doi:10.18653/v1/2024.naacl-long.219.

\bibitem{Pan.2024}
Pan S, Luo L, Wang Y, Chen C, Wang J, Wu X.
\newblock Unifying Large Language Models and Knowledge Graphs: A Roadmap.
\newblock IEEE Transactions on Knowledge and Data Engineering.
  2024;36(7):3580-99.
\newblock doi:10.1109/TKDE.2024.3352100.

\bibitem{Liang.2024}
Liang K, Meng L, Liu M, Liu Y, Tu W, Wang S, et~al.
\newblock A Survey of Knowledge Graph Reasoning on Graph Types: Static,
  Dynamic, and Multi-Modal.
\newblock IEEE transactions on pattern analysis and machine intelligence.
  2024;46(12):9456-78.
\newblock doi:10.1109/TPAMI.2024.3417451.

\bibitem{Perevalov.2022}
Perevalov A, Diefenbach D, Usbeck R, Both A.
\newblock QALD-9-plus: A Multilingual Dataset for Question Answering over
  DBpedia and Wikidata Translated by Native Speakers.
\newblock In: 16th IEEE International Conference on Semantic Computing, ICSC
  2022, Laguna Hills, CA, USA, January 26-28. IEEE; 2022. p. 229-34.
\newblock doi:10.1109/ICSC52841.2022.00045.

\bibitem{Ley.2002}
Ley M.
\newblock The DBLP Computer Science Bibliography: Evolution, Research Issues,
  Perspectives.
\newblock In: Laender AHF, Oliveira AL, editors. String Processing and
  Information Retrieval. SpringerLink B{\"u}cher. Berlin, Heidelberg:
  {Springer-Verlag Berlin Heidelberg}; 2002. p. 1-10.
\newblock doi:10.1007/3-540-45735-6{\textunderscore }1.

\bibitem{Priem.2022}
Priem J, Piwowar HA, Orr R.
\newblock OpenAlex: A fully-open index of scholarly works, authors, venues,
  institutions, and concepts.
\newblock CoRR. 2022;abs/2205.01833.
\newblock doi:10.48550/ARXIV.2205.01833.

\bibitem{Caufield.2023}
Caufield JH, Putman T, Schaper K, Unni DR, Hegde H, Callahan TJ, et~al.
\newblock KG-Hub-building and exchanging biological knowledge graphs.
\newblock Bioinformatics (Oxford, England). 2023;39(7).
\newblock doi:10.1093/bioinformatics/btad418.

\bibitem{Nourhan.2024}
Ibrahim N, Aboulela S, Ibrahim AF, Kashef RF.
\newblock A survey on augmenting knowledge graphs (KGs) with large language
  models (LLMs): models, evaluation metrics, benchmarks, and challenges.
\newblock Discov Artif Intell. 2024;4(1):76.
\newblock doi:10.1007/s 44163-024-00175-8.

\bibitem{Savas.2023}
Takan S.
\newblock Knowledge graph augmentation: consistency, immutability, reliability,
  and context.
\newblock PeerJ Comput Sci. 2023;9:e1542.
\newblock doi:10.7717/peerj-cs.1542.

\bibitem{Freitas.2020}
Freitas A, O'Riain S, Curry E.
\newblock Querying and Searching Heterogeneous Knowledge Graphs in Real-time
  Linked Dataspaces.
\newblock In: Real-time Linked Dataspaces - Enabling Data Ecosystems for
  Intelligent Systems. Springer; 2020. p. 105-24.
\newblock doi:10.1007/978-3-030-29665-0{\textunderscore }7.

\bibitem{Khan.2023}
Khan A.
\newblock Knowledge Graphs Querying.
\newblock SIGMOD Rec. 2023;52(2):18-29.
\newblock doi:10.1145/3615952.3615956.

\bibitem{Azevedo.2023}
Azevedo LG, Souza RFS, {Soares, Elton Figueiredo de Souza}, Thiago RM, Tesolin
  JCC, Oliveira AC, et~al.
\newblock A Polystore Architecture Using Knowledge Graphs to Support Queries on
  Heterogeneous Data Stores.
\newblock CoRR. 2023;abs/2308.03584.
\newblock doi:10.48550/arXiv.2308.03584.

\bibitem{Kejriwal.2022}
Kejriwal M.
\newblock Knowledge Graphs: A Practical Review of the Research Landscape.
\newblock Inf. 2022;13(4):161.
\newblock doi:10.3390/info13040161.

\bibitem{Chebotko.2009}
Chebotko A, Lu S, Fotouhi F.
\newblock Semantics preserving SPARQL-to-SQL translation.
\newblock Data Knowl Eng. 2009;68(10):973-1000.
\newblock doi:10.1016/j.datak.2009.04.001.

\bibitem{Ngomo.2013}
Ngomo ACN, B{\"u}hmann L, Unger C, Lehmann J, Gerber D.
\newblock Sorry, i don't speak SPARQL: translating SPARQL queries into natural
  language.
\newblock In: 22nd International World Wide Web Conference, WWW '13, Rio de
  Janeiro, Brazil, May 13-17, 2013; 2013. p. 977-88.
\newblock doi:10.1145/2488388.2488473.

\bibitem{Banerjee.2023}
Banerjee D, Awale S, Usbeck R, Biemann C.
\newblock DBLP-QuAD: A Question Answering Dataset over the DBLP Scholarly
  Knowledge Graph.
\newblock In: Proceedings of the 13th International Workshop on
  Bibliometric-enhanced Information Retrieval co-located with 45th European
  Conference on Information Retrieval (ECIR 2023), Dublin, Ireland, April 2nd;
  2023. p. 37-51.
\newblock Available from: \url{https://ceur-ws.org/Vol-3617/paper-05.pdf}.

\bibitem{Azmy.2018}
Azmy M, Shi P, Lin J, Ilyas IF.
\newblock Farewell Freebase: Migrating the SimpleQuestions Dataset to DBpedia.
\newblock In: Proceedings of the 27th International Conference on Computational
  Linguistics, COLING 2018, Santa Fe, New Mexico, USA, August 20-26; 2018. p.
  2093-103.
\newblock Available from: \url{https://aclanthology.org/C18-1178/}.

\bibitem{Banerjee.2022}
Banerjee D, Nair PA, Kaur JN, Usbeck R, Biemann C.
\newblock Modern Baselines for SPARQL Semantic Parsing.
\newblock In: SIGIR '22: The 45th International ACM SIGIR Conference on
  Research and Development in Information Retrieval, Madrid, Spain, July 11 -
  15; 2022. p. 2260-5.
\newblock doi:10.1145/3477495.3531841.

\bibitem{Dubey.2024}
Dubey A, Jauhri A, Pandey A, Kadian A, Al-Dahle A, Letman A, et~al.
\newblock The Llama 3 Herd of Models.
\newblock CoRR. 2024;abs/2407.21783.
\newblock doi:10.48550/arXiv.2407.21783.

\bibitem{Mistral.2024}
{Mistral AI}. Mistral Large 2. {Mistral AI}; 2024.
\newblock Accessed: May 5, 2025.
\newblock Mistral AI Blog Post.
\newblock Available from: \url{https://mistral.ai/news/mistral-large-2407}.

\bibitem{Guo.2025}
Guo D, Yang D, Zhang H, Song J, Zhang R, Xu R, et~al.
\newblock DeepSeek-R1: Incentivizing Reasoning Capability in LLMs via
  Reinforcement Learning.
\newblock CoRR. 2025;abs/2501.12948.
\newblock doi:10.48550/arXiv.2501.12948.

\bibitem{Reimers.2019}
Reimers N, Gurevych I.
\newblock Sentence-BERT: Sentence Embeddings using Siamese BERT-Networks.
\newblock In: Proceedings of the 2019 Conference on Empirical Methods in
  Natural Language Processing and the 9th International Joint Conference on
  Natural Language Processing, EMNLP-IJCNLP 2019, Hong Kong, China, November
  3-7; 2019. p. 3980-90.
\newblock doi:10.18653/v1/D19-1410.

\bibitem{Wei.2022}
Wei J, Wang X, Schuurmans D, Bosma M, ichter b, Xia F, et~al.
\newblock Chain-of-Thought Prompting Elicits Reasoning in Large Language
  Models.
\newblock In: Koyejo S, Mohamed S, Agarwal A, Belgrave D, Cho K, Oh A, editors.
  Advances in Neural Information Processing Systems. vol.~35. Curran
  Associates, Inc.; 2022. p. 24824-37.
\newblock Available from:
  \url{https://proceedings.neurips.cc/paper_files/paper/2022/file/9d5609613524ecf4f15af0f7b31abca4-Paper-Conference.pdf}.

\end{thebibliography}

\end{document}